\documentclass[11pt]{article}

\usepackage[final]{acl}

\usepackage{times}
\usepackage{latexsym}

\usepackage[T1]{fontenc}
\usepackage[utf8]{inputenc}

\usepackage{microtype}

\usepackage{inconsolata}

\usepackage{graphicx}
\graphicspath{{./figs/}}
\usepackage{array}
\usepackage{booktabs}
\usepackage{multirow}
\newcolumntype{C}[1]{>{\centering\arraybackslash}m{#1}}
\usepackage{amsmath}
\usepackage{amssymb}
\usepackage{tcolorbox}

\title{STRIDE-ED: A Strategy-Grounded Stepwise Reasoning Framework for Empathetic Dialogue Systems}

\author{
  \textbf{Hongru Ji\textsuperscript{1}},
  \textbf{Yuyin Fan\textsuperscript{1}},
  \textbf{Meng Zhao\textsuperscript{2}},
  \textbf{Xianghua Li\textsuperscript{1}}\thanks{Corresponding Author.},
  \textbf{Lianwei Wu\textsuperscript{3}} and
  \textbf{Chao Gao\textsuperscript{1}}
\\
  \textsuperscript{1}School of Artificial Intelligence, OPtics and ElectroNics (iOPEN), \\ Northwestern Polytechnical University, China
  \\
  \textsuperscript{2}School of Artificial Intelligence and Big Data, Henan University of Technology, China
  \\
  \textsuperscript{3}School of Computer Science, Northwestern Polytechnical University, China
  \\
  \{jihongru,fanyuyin\}@mail.nwpu.edu.cn,
  zm@haut.edu.cn, \\
  \{li\_xianghua,wlw,cgao\}@nwpu.edu.cn
}

\begin{document}
\maketitle
\begin{abstract}
Empathetic dialogue requires not only recognizing a user’s emotional state but also making strategy-aware, context-sensitive decisions throughout response generation.
However, the lack of a comprehensive empathy strategy framework, explicit task-aligned multi-stage reasoning, and high-quality strategy-aware data fundamentally limits existing approaches, preventing them from effectively modeling empathetic dialogue as a complex, multi-stage cognitive and decision-making process.
To address these challenges, we propose STRIDE-ED, a STRategy-grounded, Interpretable, and DEep reasoning framework that models Empathetic Dialogue through structured, strategy-conditioned reasoning.
To support effective learning, we develop a strategy-aware data refinement pipeline integrating LLM-based annotation, multi-model consistency-weighted evaluation, and dynamic sampling to construct high-quality training data aligned with empathetic strategies.
Furthermore, we adopt a two-stage training paradigm that combines supervised fine-tuning with multi-objective reinforcement learning to better align model behaviors with target emotions, empathetic strategies, and response formats.
Extensive experiments demonstrate that STRIDE-ED generalizes across diverse open-source LLMs and consistently outperforms existing methods on both automatic metrics and human evaluations.
Our data and code are publicly available.\footnote{https://github.com/jicoder-nwpu/STRIDE-ED}
\end{abstract}

\section{Introduction}
\begin{figure}[t]
	\includegraphics[width=\columnwidth]{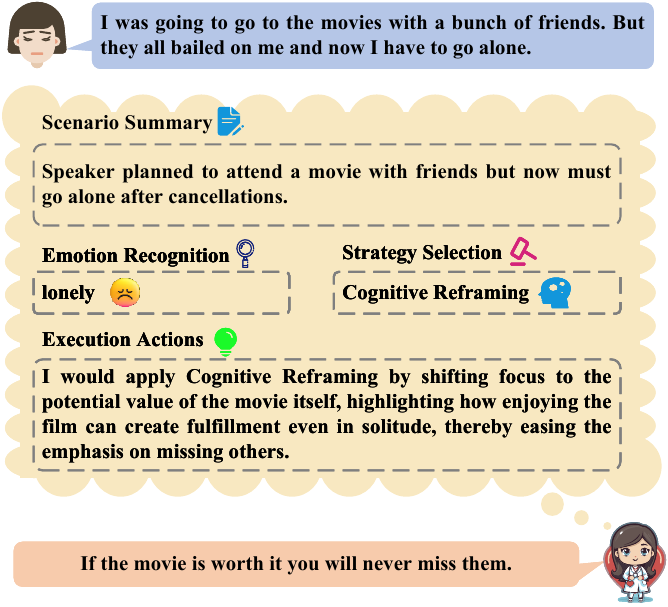}
	\caption{Illustration of the STRIDE-ED reasoning process. Given a user utterance, the framework performs scenario summarization, emotion recognition, strategy inference, and strategy-guided response generation.}
	\label{fig:intro_exam}
\end{figure}
Empathetic dialogue, a cornerstone of human social interaction, requires not only the recognition of another’s emotional state but also the formulation of a response that conveys understanding, validation, and appropriate support~\citep{batson2009these}. From a psychological perspective, this is a complex, multi-stage cognitive and decision-making process~\citep{davis1983measuring,gao2023cab}. This highlights the strategic and context-sensitive nature of empathy, making empathetic dialogue a challenging task beyond surface-level text generation.

Early studies focused on implicitly enhancing dialogue models through external commonsense knowledge or affective lexicons. For example, prior works~\citep{liu2022empathetic,cai2023improving} incorporates emotional commonsense graphs or commonsense knowledge selection into neural architectures to provide latent cues for more emotion-aware generation. However, without explicit modeling of reasoning or decision processes~\citep{zhang2026semanticawarelogicalreasoningsemiotic, zhang2026logicalphasetransitionsunderstanding, chen2025red, luo2026gcotdecodingunlockingdeepreasoning, wang2026one}, these approaches offer limited insight into the mechanisms through which emotional understanding informs response generation.

To address this lack of transparency, recent research has turned to Large Language Models (LLMs), which enable more explicit modeling of intermediate reasoning processes through Chain-of-Thought (CoT) prompting. Building on this direction, prior work such as \citep{hu2025beyond} further encourages LLMs to generate explicit intermediate rationales before producing responses, thereby enhancing interpretability~\citep{zhang2024escot, chen2025improving}.
However, a fundamental limitation persists in these CoT-based approaches: the reasoning process lacks grounding in a rigorous strategic framework. Existing strategies~\citep{liu2021towards, zhangetal2025intentionesc} often derived from specific domains such as psychological support, primarily address negative emotions and are confined to low-level responses.
Emotions in dialogue typically span positive, negative, and neutral states~\citep{ji2025hybrid}, yet existing methods fail to fully capture this spectrum and lack support for higher-order cognitive strategies, as shown in Figure~\ref{fig:intro_exam}.
Consequently, while these existing work structured in form, the model’s reasoning steps frequently remain superficial and exhibit inconsistent strategic coherence.

In summary, existing approaches suffer from three key limitations.
(1) \textbf{Incomplete Empathy Strategy Coverage.} Prior methods lack a comprehensive strategy system covering diverse emotional states and higher-order cognition, limiting principled decision-making.
(2) \textbf{Lack of Task-Aligned Multi-Stage Reasoning.} CoT-based methods do not explicitly model dialogue as a task-specific multi-stage reasoning process.
(3) \textbf{Insufficient Strategy-Aware Supervision.} Training data lack sufficient high-quality annotations aligned with empathetic strategies and reasoning.

To address these issues, we propose STRIDE-ED, a framework that proceeds by first establishing a comprehensive strategy system, which then enables task-aligned reasoning through a dedicated data pipeline. At its core, we construct a unified strategy system covering positive, neutral, and negative emotions to guide response generation. Leveraging this system, we automatically annotate the EMPATHETICDIALOGUES dataset \citep{rashkin2019towards} with strategy types and rationales using authoritative LLMs. Subsequently, a rigorous data refinement process employs multi-LLM evaluation with consistency-weighted scoring and strategy-aware sampling to curate high-quality training subsets. Finally, the model is optimized through a two-stage training paradigm that combines supervised fine-tuning with multi-objective reinforcement learning, ensuring alignment with strategic, emotional, and structural correctness.
Our contributions can be summarized as follows:

\begin{itemize}
	\item We propose an interpretable framework for empathetic dialogue, STRIDE-ED, featuring a comprehensive empathy strategy system and stepwise decision-making.
	
	\item To support effective training, a strategy-aware data refinement pipeline is constructed, combining LLM-based annotation, multi-model weighted evaluation, and dynamic sampling to regulate strategy distribution and difficulty.
	
	\item A two-stage training paradigm is introduced, in which supervised fine-tuning establishes reasoning, followed by reinforcement learning to improve emotional alignment, strategy execution, and response consistency.
	
	\item Extensive experiments show that our framework generalizes across diverse open-source LLMs and achieves superior performance on both automatic and human evaluations.
\end{itemize}

\section{Related Works}

\subsection{Implicit Knowledge-Driven Empathy}
Early empathetic dialogue models lacked sufficient prior knowledge, limiting their understanding of users' emotions and contexts. To address this, some studies incorporate external knowledge to broaden the model's perspective~\citep{zhang2026stable} and enhance reasoning.
For example, \citet{ghosal2020cosmic} leveraged structured commonsense knowledge from ATOMIC~\citep{sap2019atomic}, such as mental states and causal relations, to model interlocutor interactions for improved emotion understanding.
\citet{zhong2021care} integrated commonsense-aware emotional latent concepts to generate emotionally appropriate responses, while \citet{sabour2022cem} inferred users’ situational contexts through commonsense-based reasoning.
Further, \citet{cai2023improving} introduced an adaptive commonsense knowledge selection mechanism to refine contextual cognition, and \citet{qiao2025multi} constructed multi-hop reasoning graphs to incorporate external knowledge during response generation.
However, these methods did not explicitly model strategy-guided decision-making.
\subsection{Explicit Reasoning with CoT for Empathy}
LLMs possessed extensive knowledge reserves, and CoT prompting~\citep{wei2022chain} enabled stepwise reasoning, facilitating more explicit decision-making in empathetic dialogue.
Building on this paradigm, \citet{tu2022misc} inferred fine-grained user emotions and employed a mixed strategy mechanism for response generation, while \citet{chen2023dynamic} dynamically generated counseling strategies via zero-shot prompting to guide personalized responses.
Subsequent work further enhanced empathy through data and structural designs: \citet{chen2023soulchat} fine-tuned LLMs on consultant-style multi-turn dialogues, and \citet{ye2025sweetiechat} adopted a strategy-enhanced role-playing framework with multiple interacting roles to generate diverse training data.
Finally, \citet{zhangetal2025intentionesc} introduced an intention-centered framework that mapped inferred supporter intentions to support strategies using a chain-of-thought mechanism.
Despite improved interpretability, these CoT-based approaches lacked comprehensive strategy coverage and did not explicitly model the full, structured reasoning process underlying human empathetic decision-making.

\section{Methodology}
\begin{figure*}[t]
	\includegraphics[width=\textwidth]{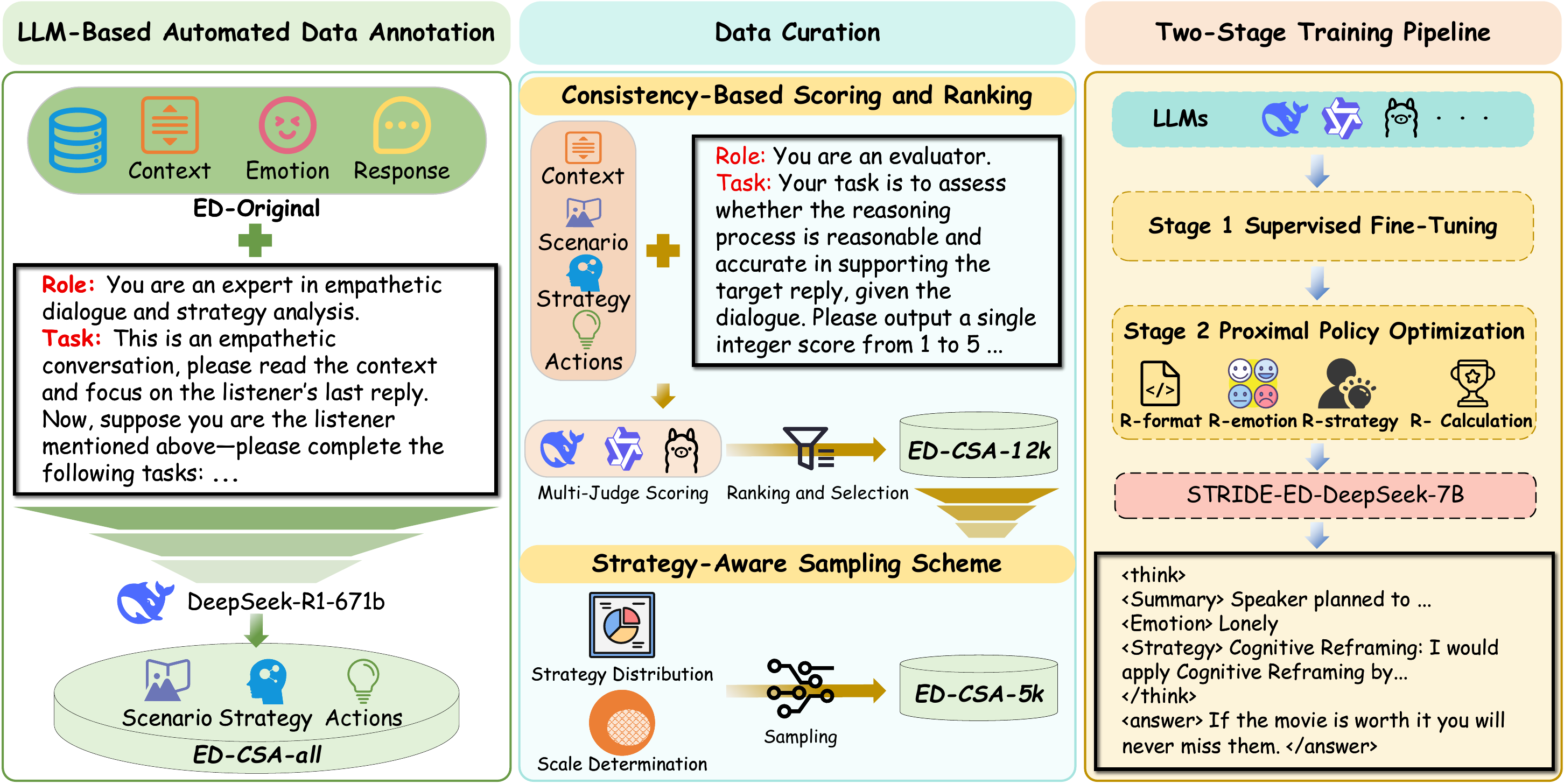}
	\caption{The architecture of the STRIDE-ED framework, illustrating the complete pipeline from data preparation and refinement to model training.}
	\label{fig:framework}
\end{figure*}

STRIDE-ED is a general-purpose framework for empathetic dialogue applicable to diverse open-source LLMs. It implements a comprehensive empathy strategy system and a task-aligned, multi-step CoT paradigm to model dialogue as a progressive cognitive and decision-making process. At the data and training levels, STRIDE-ED integrates LLM-based automatic annotation, strategy-aware data refinement, and two-stage optimization. An overview is shown in Figure~\ref{fig:framework}.

\subsection{Task Formulation}
Empathetic dialogue consists of a multi-turn interaction between a user and a conversational agent.
We denote the dialogue history as $\mathcal{C} = \{u_1, u_2, \ldots, u_{t-1}\}$, where $u_i$ represents the utterance at the $i$-th turn.
Each utterance $u_i$ is a sequence of tokens, i.e., $u_i = (w_{i,1}, w_{i,2}, \ldots, w_{i,n_i})$.
The goal of empathetic dialogue is to generate an empathetic response $u_t$ while appropriately recognizing the user’s emotional state $e$.
In this work, we introduce auxiliary objectives and model empathetic dialogue as a sequential generation process.
Specifically, conditioned on $\mathcal{C}$, the model generates a dialogue scenario summary $\textit{sum}$, infers the target emotional state $e$, determines an empathy strategy $\textit{stra}$ along with its execution actions $\textit{acts}$, and finally produces the empathetic response $u_t$, corresponding to modeling the conditional distribution $P(\textit{sum}, e, \textit{stra}, \textit{acts}, u_t \mid \mathcal{C})$.

\subsection{Empathy Strategy System}
In empathetic dialogue modeling, response strategies form a crucial intermediate stage between emotion understanding and response generation. Rather than generating surface-level replies directly, effective models must explicitly decide how to respond based on the user’s emotional state and dialogue context. \citet{liu2021towards} proposed a widely adopted taxonomy of eight empathy strategies, primarily applied in counseling-oriented settings that focus on negative emotions. Higher-order cognitive strategies, however, remain underexplored, limiting the effectiveness of current systems in complex dialogues requiring sophisticated reasoning.

Motivated by these observations, we first analyze the emotional distributions in the EMPATHETICDIALOGUES dataset and conduct a fine-grained examination of response content. Based on these analyses, we expand the original empathy strategy set to better accommodate positive, neutral, and negative emotional contexts, and assign a three-level difficulty rating (I–III) to each strategy to reflect the cognitive complexity involved in its application. In particular, the original Question strategy is further subdivided into three distinct \emph{Exploring}-type strategies to capture more nuanced cognitive and emotional interactions. The structured strategy system guides both annotation and model training (full details are provided in Appendix~\ref{sec:appendix_strategy}).

\subsection{Stepwise Cognitive CoT Design}
Inspired by insights from cognitive psychology, we model the human thought process during empathetic dialogue as a stepwise, incremental deliberation. Upon receiving a speaker’s utterance, individuals first infer missing information or make educated guesses about the described situation based on prior knowledge and contextual understanding, forming a high-level situational representation. This situational comprehension allows them to adopt the speaker’s perspective and facilitates accurate recognition of the speaker’s emotional state.
Next, humans deliberate on which response strategy to employ, guided by the speaker’s emotional state and contextual cues. Because strategies represent generalized methods, their concrete execution may differ across scenarios. To account for this variability, we introduce an action inference stage that bridges strategy selection and the generation of the final response.

This stepwise CoT design captures the structured cognitive progression in human empathetic reasoning, providing a principled framework for multi-stage, strategy-aware response generation. Implementation leverages structured intermediate tags—such as <Context>, <Emotion>, and <Strategy>—to guide the model’s internal reasoning, culminating in the final response.

\subsection{LLM-Based Automated Data Annotation}
Building upon the aforementioned theoretical framework, we address the absence of explicit annotations required for model training. Specifically, the adopted dataset lacks annotations for dialogue scenario summaries, empathy strategy types, and the concrete actions used to implement each strategy. To this end, we employ DeepSeek-R1\footnote{https://huggingface.co/deepseek-ai/DeepSeek-R1}, a large language model with strong reasoning capabilities, as an automated annotation expert. Carefully designed annotation prompts are used to generate structured labels for each dialogue instance, as shown in Appendix~\ref{sec:appendix_prompts}. The annotated dataset resulting from this process is denoted as \textit{ED-CSA-all}.

\subsection{Consistency-Based Scoring and Ranking}
To ensure the reliability of the automatically annotated data while mitigating the high cost of large-scale manual verification, we employ a multi-judge evaluation mechanism based on LLMs. Specifically, three representative models (DeepSeek-R1, Qwen3\footnote{https://huggingface.co/Qwen/Qwen3-8B}, and Llama-3.1\footnote{https://huggingface.co/meta-llama/Llama-3.1-8B-Instruct}) are selected as independent expert evaluators. Using carefully designed evaluation prompts, as shown in Appendix~\ref{sec:appendix_prompts}, each model assesses every example in the \textit{ED-CSA-all} dataset and assigns a quality score ranging from 1 to 5, where higher scores indicate stronger semantic coherence and alignment among the annotated scenario summary, selected strategy, inferred action, and the corresponding dialogue context. Models are restricted to generate integer scores only.

We apply a reliability-weighted multi-judge aggregation framework to fairly integrate scores from multiple LLM evaluators.  
Let $\mathcal{M} = \{m_1, m_2, m_3\}$ denote the set of evaluators, corresponding to DeepSeek-R1, Qwen, and LLaMA. For each $m_i \in \mathcal{M}$, let $\mathbf{s}_i$ denote its score vector over the dataset. The reliability of each evaluator is estimated as the average Spearman correlation~\citep{spearman1904proof} with the others:
\begin{equation}
	\rho_i = \frac{1}{2} \sum_{j \neq i} \mathrm{Spearman}(\mathbf{s}_i, \mathbf{s}_j), \quad \forall\, i \in \mathcal{M}.
\end{equation}

The final quality score for a sample $x$ is computed as a reliability-weighted sum of evaluator scores with an agreement-based penalty:
\begin{equation}
	S(x) = \sum_{i=1}^{|\mathcal{M}|} 
	\frac{\rho_i}{\sum_{k=1}^{|\mathcal{M}|} \rho_k} \, s_i(x)
	- \lambda \, \sigma(x),
\end{equation}
where $s_i(x)$ denotes the score assigned to sample $x$ by evaluator $m_i$, and $\sigma(x)$ represents the standard deviation of the scores $\{s_i(x)\}_{i=1}^{|\mathcal{M}|}$ across evaluators, measuring inter-rater disagreement.  
The hyperparameter $\lambda$ is set to $0.1$ in our experiments. This formulation favors samples with both high weighted scores and strong evaluator consensus.

Based on the resulting score ranking, we select the top 12k samples as the candidate pool for subsequent sampling, denoted as \textit{ED-CSA-12k}.

\begin{figure}[t]
	\includegraphics[width=\columnwidth]{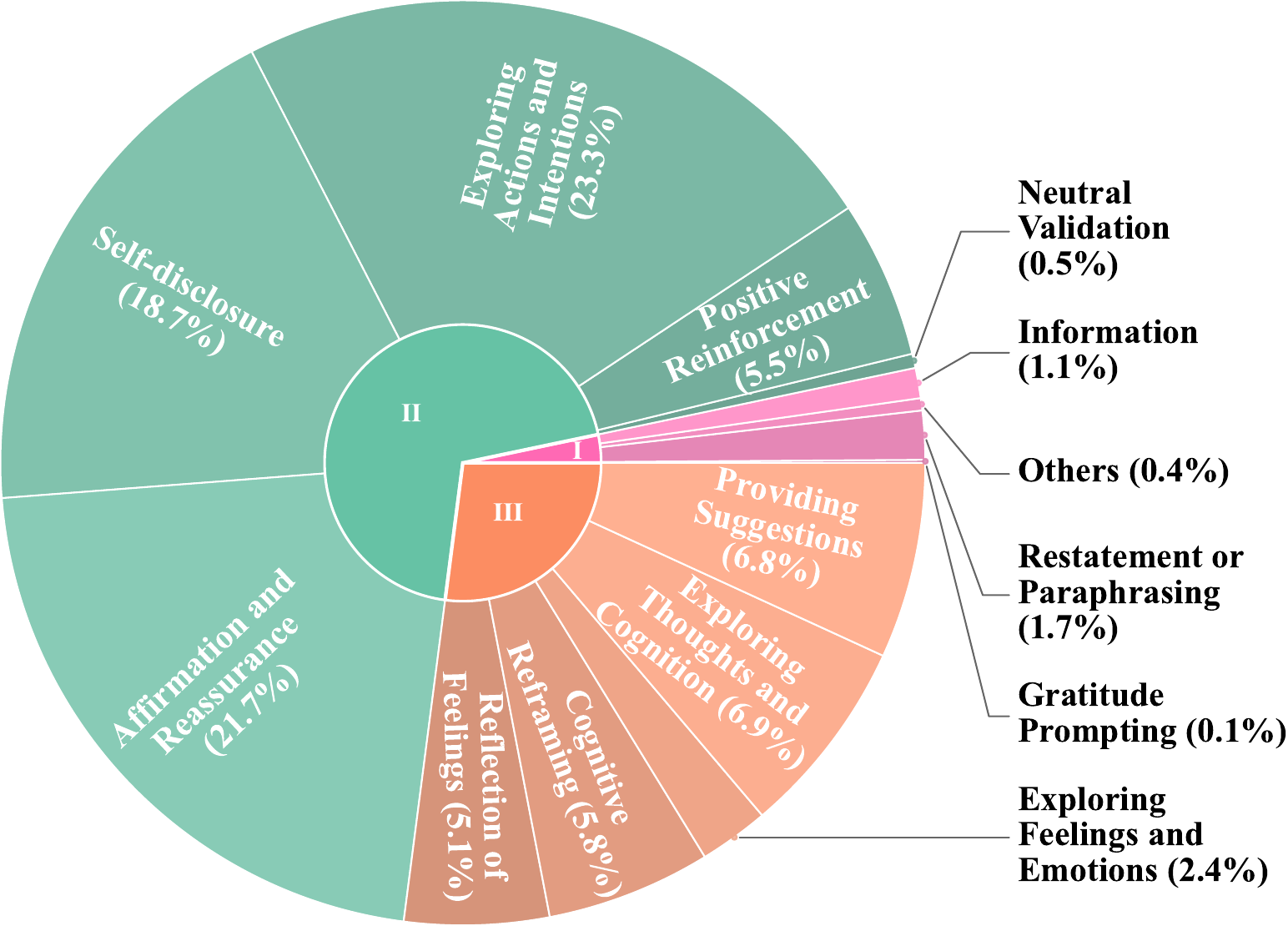}
	\caption{Distribution of Strategy Types in \textit{ED-CSA-all}.}
	\label{fig:strategy_pie}
\end{figure}

\subsection{Strategy-Aware Sampling Scheme}
We further emphasize that strategy usage should reflect both distributional characteristics and contextual sensitivity rather than being applied uniformly. The distribution of strategy types in \textit{ED-CSA-all} is illustrated in Figure~\ref{fig:strategy_pie}. Data sorting and filtering inherently reshape the empirical strategy distribution, and directly training on such processed data may distort natural response patterns. Simultaneously, the model should learn selective strategy reasoning based on dialogue context, reserving explicit strategy and action deliberation for scenarios that demand higher cognitive effort.

Based on the above considerations, we design a strategy-aware sampling scheme in which the target sampling distribution is jointly determined by the empirical frequency of each strategy and its associated difficulty level.
Let $\mathcal{S}=\{s_1,\dots,s_{|\mathcal{S}|}\}$ denote the set of empathy strategies.
We first compute the empirical frequency of each strategy $s_i$ in the full annotated dataset \textit{ED-CSA-all}, denoted as $a_i$.
Next, we assign a difficulty weight $d_i$ to each strategy according to its difficulty level, where higher-level strategies receive larger weights.
We then compute a weighted score for each strategy by combining its empirical frequency with its assigned difficulty level, and normalize these scores to obtain the target sampling proportions:
\begin{equation}
	p_i = \frac{a_i \cdot d_i}{\sum_{j=1}^{|\mathcal{S}|} a_j \cdot d_j}.
\end{equation}

This distribution favors the retention of infrequent yet cognitively demanding strategies, while preventing low-difficulty strategies from disproportionately dominating the sampled data.
Subsequently, given the ranked candidate pool \textit{ED-CSA-12k}, we apply a binary search procedure to determine the maximum subset size whose empirical strategy distribution can satisfy the target proportions ${p_i}$. Based on this subset size and the target distribution, we compute the required number of samples for each strategy and perform stratified random sampling to construct the refined training set \textit{ED-CSA-5k}.

\subsection{Model Training}
After constructing the refined training data, inspired by prior training methods~\citep{liu2026profit, limol, li2026arise, xue2026supervisedfinetuningfailslearn, xue2026reasonneededefficientgenerative}, we adopt a two-stage optimization paradigm.

\textbf{Supervised Fine-Tuning.}
In the first stage, we construct the supervised training set by combining the curated subset \textit{ED-CSA-5k} with the remaining samples from \textit{ED-CSA-all}, for which strategy annotations are removed. This design ensures that only the curated subset provides explicit strategy guidance, while the remaining samples contribute to general response generation without strategy supervision. The base LLM $\mathcal{M}$ is then fine-tuned to sequentially generate both the intermediate reasoning components and the final response, yielding the model denoted as STRIDE-ED-$\mathcal{M}$-SFT. The supervised training objective minimizes the negative log-likelihood over the structured output sequence:
\begin{equation}
	\mathcal{L}_{\mathrm{SFT}}
	= - \frac{1}{N} \sum_{n=1}^{N}
	\log p_\theta(y \mid \mathcal{C}),
\end{equation}
where $y$ represents the target output sequence conditioned on the dialogue history $\mathcal{C}$.

\textbf{Reinforcement Learning.}
In the second stage, we further refine the SFT model via reinforcement learning on \textit{ED-CSA-12k}. Beginning with STRIDE-ED-$\mathcal{M}$-SFT, we apply Proximal Policy Optimization (PPO) to optimize the model towards outputs that better adhere to the desired format, emotional alignment, and strategy execution. The reward function is defined as follows:
\begin{align}
	r_{\text{format}} &= 
	\begin{cases}
		1, & \text{if format is correct} \\
		0, & \text{otherwise}
		\label{eq:format}
	\end{cases} \\
	r_{\text{emotion}} &= 
	\begin{cases}
		1, & \text{if emotion is correct} \\
		0, & \text{otherwise}
	\end{cases} \\
	r_{\text{strategy}} &= 
	\begin{cases}
		0, & \text{if strategy is incorrect} \\
		1, & \text{otherwise}
	\end{cases} \\
	R &= r_{\text{format}} \cdot \Big( 1 + r_{\text{emotion}} + r_{\text{strategy}} \Big)
\end{align}
where $r_{\text{format}}$, $r_{\text{emotion}}$, and $r_{\text{strategy}}$ represent the reward components corresponding to format compliance, emotional accuracy, and strategic appropriateness, respectively. The format constraint is specified as “<think>.</think><answer>.</answer>”, and a regular expression is employed to verify whether the generated output adheres to this structural requirement. The overall reward $R$ is designed to prioritize format correctness while jointly encouraging accurate emotional expression and effective strategic reasoning.
The resulting model is denoted as STRIDE-ED-$\mathcal{M}$.

\section{Experiments}
\subsection{Datasets}
We adopt the EMPATHETICDIALOGUES dataset~\citep{rashkin2019towards} and the ESConv dataset~\citep{liu2021towards} for our experiments. Developed by Facebook AI Research, EMPATHETICDIALOGUES is a large-scale multi-turn dialogue corpus dedicated to enhancing the empathetic capabilities of conversational systems. It comprises approximately 25,000 dialogue scenarios grounded in specific emotional contexts, covering 32 emotion labels. In the dataset, participants are assigned the roles of \textit{speaker} and \textit{listener} to simulate human interactions around emotional experiences. 
ESConv is a multi-turn dialogue dataset for emotional support conversations, which focuses on modeling support strategies.
\subsection{Baselines}
To ensure a comprehensive evaluation, we select baseline models spanning three major research trajectories in empathetic dialogue generation:
(1) Traditional architectures that focus on neural network design for emotional modeling, including Transformer~\citep{vaswani2017attention}, MoEL~\citep{lin2019moel}, EmpDG~\citep{li2020empdg} and GLHG~\cite{peng2022control}.
(2) External-knowledge-enhanced methods that incorporate commonsense or causal graphs to enrich responses, such as KEMP~\citep{li2022knowledge}, DCKS~\citep{cai2023improving}, E-CORE~\citep{fu2023core}, Emp-USIR~\citep{jiang2023emp} and EESC~\cite{hao2025enhancing}.
(3) Reflective decision-integration approaches that explicitly model internal cognitive processes for strategic response generation, represented by CAB~\citep{gao2023cab}, IAMM~\citep{yang2024iterative}, Emstremo~\cite{li2024emstremo}, Sibyl~\cite{wang-etal-2025-sibyl}, and ReflectDiffu~\citep{yuan2025reflectdiffu}.
\subsection{Implementation Details}
We implement all LLM training using the verl~\footnote{https://github.com/volcengine/verl} framework, with DeepSeek-7B-Chat selected as the primary model.\footnote{https://huggingface.co/deepseek-ai/deepseek-llm-7b-chat}
During the supervised fine-tuning stage, the initial learning rate is set to $1 \times 10^{-4}$, the batch size is 16, and the maximum input sequence length is capped at 2048 tokens.
In the reinforcement learning stage, the total batch size is increased to 128, with the maximum generation length limited to 1024 tokens.
We follow a fixed 8:1:1 split for training, validation, and testing.
All experiments are conducted on four NVIDIA A800 GPUs, each equipped with 80 GB of memory.
\begin{table*}
	\centering
	\renewcommand{\arraystretch}{1}
	\begin{tabular}{lccccccccc}
		\toprule
		\textbf{Models} & \textbf{B-1$\uparrow$} & \textbf{B-2$\uparrow$} & \textbf{B-3$\uparrow$} & \textbf{B-4$\uparrow$} & $\textbf{Acc}_{emo}\uparrow$ &\textbf{D-1$\uparrow$} & \textbf{D-2$\uparrow$} & \textbf{BARTScore$\uparrow$} & \textbf{PPL$\downarrow$} \\
		\midrule
		Transformer & 18.07 & 8.34 & 4.57 & 2.86 & -- & 0.36 & 1.35 & -  & 37.62 \\
		MoEL       & 18.02 & 8.67 & 4.35 & 2.73 &  31.02 & 0.43 &  1.76 & 0.56  & 36.81 \\
		EmpDG       & 19.96 & 9.11 & 4.74 & 2.80 & 31.65 & 0.46 & 1.99 & \underline{0.57}  & 37.43 \\
		KEMP        & 18.07 & 8.30 & 4.37 & 2.65 & 36.40 & 0.55 & 2.29 & 0.52  &  36.89 \\
		DCKS        & 21.73 & 10.62 & \underline{6.24} &\underline{4.09} & --& 2.19 & 9.61 & -  & \underline{16.08} \\
		E-CORE     & 19.77 & 5.65 & 3.28 & 2.11 & --& 0.68 & 3.38 & -  & 33.04 \\
		Emp-USIR    & 20.11 & 9.86 &5.72 & 3.73 & --& 0.66 & 3.10 & - & 35.29 \\ 
		CAB        &  20.23 &  9.39 &   4.96 &  3.01 & 40.52&   0.89 & 2.95 & - &  35.06 \\ 
		IAMM        & 19.51 & 8.74 &  4.86 & 3.32 & 43.72&  0.88 & 3.05 & - & 25.94 \\ 
		Sibyl & 21.45 & 9.35 & 5.01 & 2.95 & - & \textbf{5.65} & \textbf{36.11} & - & - \\
		ReflectDiffu &\underline{23.59} &\underline{11.25} & 5.35 & 3.62 &\underline{48.76} & 0.98 & 4.35 & 0.56 & 24.56 \\ \hline
		Ours &\textbf{24.54}	&\textbf{11.96}	&\textbf{7.13}	&\textbf{4.67}	&\textbf{57.25}	&\underline{2.56}	&\underline{13.63} &\textbf{0.71}	&\textbf{10.50} \\
		\bottomrule
	\end{tabular}
	\caption{\label{tab:exp_main}
		Automatic evaluation results on the EMPATHETICDIALOGUES dataset. Boldface and underline indicate the best and second-best values, respectively. "Ours" refers to STRIDE-ED-DeepSeek-7B.
	}
\end{table*}

\subsection{Evaluation Metrics}
To comprehensively evaluate STRIDE-ED, we conduct both automatic and human assessments.

\textbf{Automatic Evaluation.}
We evaluate response quality using Perplexity (PPL), BLEU (B-n), Distinct (D-n), BARTScore, and emotion accuracy (Acc$_{emo}$). PPL measures fluency, with lower values indicating greater coherence. BLEU captures relevance to reference responses via overlap, while Distinct reflects lexical diversity. BARTScore measures semantic similarity between generated and reference responses~\citep{liu2025learning}, providing a proxy for empathy. Acc$_{emo}$ measures the consistency between predicted and true emotions.

\textbf{Human Evaluation.}
We perform A/B testing to assess Empathy, Relevance, and Fluency. A panel of three annotators is employed, and 1000 dialogue turns generated by STRIDE-ED and baselines are selected for comparison, ensuring the consistency and interpretability of the evaluation outcomes.

\begin{table}[htbp]
	\centering
	\begin{tabular}{p{0.123\textwidth}C{0.05\textwidth}   % 序号列
			C{0.08\textwidth}    % Strategy
			C{0.05\textwidth}    % Definition
			C{0.05\textwidth}}
		\toprule
		\textbf{Models} & \textbf{B-1$\uparrow$} & $\textbf{D-1}\uparrow$ & \textbf{D-2$\uparrow$} & \textbf{PPL$\downarrow$} \\
		\midrule
		Transformer & - & 1.29 &	6.91 &	81.55\\
		MoEL &	19.04 & 2.33 &	15.26 &	62.93 \\
		GLHG &	19.66 & 3.50 &	21.61 &	15.67 \\
		Emstremo &	20.96 & 2.90 &	14.80 &	- \\
		EESC &	\underline{21.38} &	\underline{4.88} &	\underline{25.95} &	\underline{14.88} \\
		\hline
		Ours &\textbf{22.03}	&\textbf{5.11} &\textbf{29.63} &\textbf{10.09} \\ 
		\bottomrule
	\end{tabular}
	\caption{\label{tab:exp_esc}
		Automatic evaluation results on the ESConv.}
\end{table}
\section{Results and Discussion}
\textbf{Automatic Evaluation Results.}
As shown in Table~\ref{tab:exp_main} and Table~\ref{tab:exp_esc}, we conduct five independent runs on each dataset and report the mean results. Our method consistently outperforms all baselines across automatic evaluation metrics.
Compared with the strongest baseline, ReflectDiffu, our method achieves consistent improvements across all BLEU scores, with relative gains of 4.0\% on BLEU-1, 6.3\% on BLEU-2, 33.3\% on BLEU-3, and 29.0\% on BLEU-4. On ESConv, it also attains the highest BLEU-1, surpassing EESC. In terms of response quality, our method achieves the highest BARTScore, significantly outperforming all baselines, indicating better semantic alignment with references and enhanced empathetic quality. For emotional controllability, it improves emotion accuracy by 17.4\% over ReflectDiffu in $\mathrm{Acc}_{emo}$.
Our method also demonstrates stronger lexical diversity. On the main dataset, it achieves high Distinct-1 and Distinct-2 scores; on ESConv, it obtains the best Distinct-1 and Distinct-2, with relative improvements of 4.7\% and 14.2\% over EESC, respectively. In addition, it achieves lower perplexity, reflecting improved fluency and generation confidence.
Overall, these results show that our method not only surpasses existing empathetic dialogue models but also achieves a better balance among relevance, diversity, and fluency, validating the effectiveness of its structured reasoning and training framework.

\begin{table}[htbp]
	\centering
	\begin{tabular}{p{0.15\textwidth}
			C{0.055\textwidth}
			C{0.045\textwidth}
			C{0.045\textwidth}
			C{0.045\textwidth}}
		\toprule
		Comparison & Aspects & Win & Lose & Tie \\
		\hline										
		\multirow{3}{*}{Ours vs. MOEL} & Emp. & \textbf{46.1} & 23.6 & 30.3 \\
		& Rel. & \textbf{39.8} & 22.4 & 37.8 \\
		& Flu. & \textbf{33.5} & 15.6 & 50.9 \\
		\hline
		\multirow{3}{*}{Ours vs.  EmpDG} & Emp. & \textbf{51.6} & 18.5 & 29.9 \\
		& Rel. & \textbf{50.3} & 15.3 & 34.4 \\
		& Flu. & \textbf{35.1} & 13.5 & 51.4 \\\hline
		\multirow{3}{*}{Ours vs.  CAB} & Emp. & \textbf{53.2} & 23.2 & 23.6 \\
		& Rel. & \textbf{56.6} & 21.4 & 22.0 \\
		& Flu. & \textbf{31.5} & 11.6 & 56.9 \\
		\bottomrule
	\end{tabular}
	\caption{\label{tab:exp_human}
		Human A/B evaluation results on the EMPATHETICDIALOGUES dataset.
	}
\end{table}
\noindent \textbf{Human Evaluation Results.}
Table~\ref{tab:exp_human} presents human A/B testing results for STRIDE-ED and several baseline models on the EMPATHETICDIALOGUES dataset, evaluated in terms of empathy (Emp.), relevance (Rel.), and fluency (Flu.). The results highlight the performance characteristics of each model in generating empathetic, coherent, and contextually relevant responses.

\begin{table}[htbp]
	\centering
	\begin{tabular}{p{0.123\textwidth}C{0.05\textwidth}   % 序号列
			C{0.08\textwidth}    % Strategy
			C{0.05\textwidth}    % Definition
			C{0.05\textwidth}}
		\toprule
		\textbf{Models} & \textbf{B-1$\uparrow$} & $\textbf{Acc}_{emo}\uparrow$ & \textbf{D-2$\uparrow$} & \textbf{PPL$\downarrow$} \\
		\midrule
		Ours &24.66	&57.57	&13.68 &9.26 \\ \hline
		w/o $e$ &23.58	&- &13.66 &10.47 \\
		w/o $sum$ &22.91	&54.14 &13.42  &7.78 \\
		w/o $stra$ &22.44	&54.58 &15.09  &6.98 \\
		w/o CoT &22.55	&- &13.26  &8.64 \\ \hline
		w/o $R.\&S.$ &22.22	&55.93 &15.25  &11.86 \\
		w/o $S.$ &23.44	&56.05 &14.40  &9.89 \\ \hline
		w/o PPO &23.52	&54.48 &14.76  & 2.03\\ 
		\bottomrule
	\end{tabular}
	\caption{\label{tab:exp_ablation}
		Evaluation results of the ablation study on the EMPATHETICDIALOGUES dataset. $R.\&S.$ refer to the two modules: Consistency-Based Scoring and Ranking, and Strategy-Aware Sampling Scheme.}
\end{table}
\noindent \textbf{Ablation Study.}
Table~\ref{tab:exp_ablation} reports the ablation results of STRIDE-ED, covering both the stepwise cognitive reasoning components, data curation strategies, and PPO training. Overall, removing any component leads to performance degradation in one or more dimensions, confirming each module contributes non-trivially to the final model behavior.

\noindent \textbf{\textit{Stepwise Cognitive CoT Design.}}
Removing emotion reasoning (\textbf{w/o $e$}) causes a clear drop in BLEU-1 and an increase in PPL, indicating that explicit emotion modeling not only improves emotional alignment but also stabilizes generation quality.
Eliminating scenario summarization (\textbf{w/o $sum$}) results in further degradation in BLEU-1 and emotion accuracy, suggesting that concise situation abstraction is crucial for grounding subsequent emotional and strategic reasoning.
When strategy reasoning is removed (\textbf{w/o $stra$}), emotion accuracy drops and fluency degrades, while diversity increases.  This suggests that without strategy constraints, the model generates more varied but less controlled responses.  Overall, strategy reasoning serves as a critical coordinating component in the framework, helping balance emotional controllability, coherence, and expressive diversity.
Notably, removing structured reasoning (\textbf{w/o} CoT) leads to a drop in BLEU-1 while slightly reducing perplexity, suggesting that although responses become marginally easier to generate, they lose contextual relevance and cognitive grounding. This underscores the role of the stepwise cognitive CoT in enhancing response quality through structured reasoning rather than merely improving fluency.

\noindent \textbf{\textit{Data Scoring and Sampling.}}
Removing both consistency-based scoring and strategy-aware sampling (\textbf{w/o} $R.\&S.$) leads to the most severe degradation in BLEU-1, emotion accuracy, and perplexity, underscoring the role of systematic data filtering and distribution-aware sampling. Notably, while Dist-2 improves slightly, this likely reflects increased superficial diversity at the cost of coherence and relevance, as indicated by declines in other metrics.
When strategy-aware sampling is replaced with random sampling at the same scale (\textbf{w/o} $S.$), performance still drops compared to the full model. This confirms a balanced, difficulty-aware strategy distribution is essential, and naive random sampling can distort strategic learning.

\noindent \textbf{\textit{PPO Training.}}
Removing the PPO stage results in a pronounced drop in both perplexity and emotion accuracy, reflecting a regression to generic and uncalibrated outputs. This highlights that PPO is essential for aligning the model along key dimensions, including proper formatting, strategy adherence, and emotional fidelity, extending beyond mere instruction following.

\begin{table}[htbp]
	\centering
	\begin{tabular}{p{0.1397\textwidth}C{0.0495\textwidth}   % 序号列
			C{0.065\textwidth}    % Strategy
			C{0.0498\textwidth}    % Definition
			C{0.049\textwidth}}
		\toprule
		\textbf{Models} & \textbf{B-1$\uparrow$} & $\textbf{Acc}_{emo}\uparrow$ & \textbf{D-2$\uparrow$} & \textbf{PPL$\downarrow$} \\
		\midrule
		Ours      &24.66	&57.57	&13.68 &9.26 \\ \hline
		1/2 Data  &23.43	&56.35  &15.07 &12.75 \\
		1/4 Data  &23.37	&55.12  &14.56 &8.80 \\
		1/8 Data  &23.28	&51.29  &14.50 &9.70 \\
		1/16 Data &3.32	   	&45.53  &3.38  &6.21 \\
		Zero Data &14.86	&--     &21.49 &1.09 \\ \hline
		Qwen3-0.6B           &21.29	&51.00 &10.46  &9.79 \\
		Qwen3-4B             &20.49	&56.35 &12.51  &10.20 \\
		Qwen3-4B-In.         &22.76	&55.93 &14.32  &7.87 \\
		LLama3.2-3B            &22.77	&57.06 &14.49  &11.11 \\
		GLM-Z1-9B          &22.14	&56.98 &14.34   &10.51 \\
		\bottomrule
	\end{tabular}
	\caption{\label{tab:exp_rob}
		Evaluation results for training set size and backbone model analyses. The abbreviation "Qwen3-4B-In." denotes Qwen3-4B-Instruct.
	}
\end{table}
\noindent \textbf{Training Set Size Analysis.}
Table~\ref{tab:exp_rob} shows the results of training data size analysis on \textit{ED-CSA-all}. Model competence remains stable until data is reduced to 1/8, beyond which performance collapses sharply at 1/16—indicating severe overfitting and loss of task capability. The zero-data setup shows that while the pretrained model retains strong generative fluency, it cannot adapt to the task without in-domain training. Overall, sufficient data is essential for effective adaptation, with a clear minimum required to maintain robust performance.

\noindent \textbf{Backbone Analysis.}
Our method exhibits broad applicability and competitive performance across diverse LLM architectures and scales, as shown in Table~\ref{tab:exp_rob}. In the Qwen family, emotion accuracy improves with model size, whereas BLEU-1 does not scale monotonically—showing that compact models can match or surpass larger ones in fluency with our approach. Across Qwen, LLama, and GLM families, the framework delivers consistently strong results: LLama3.2-3B leads in emotion accuracy and diversity, while Qwen3-4B-Instruct excels in BLEU-1 and fluency. Details of the backbone LLMs are provided in Appendix~\ref{sec:appendix_exp_details}. These findings confirm that our framework is architecture-agnostic and parameter-efficient, delivering robust performance across different LLMs and scales.

\noindent \textbf{Case Study}
Representative case studies are presented to illustrate the reasoning process and interpretability of our framework, with details provided in Appendix~\ref{sec:appendix_case_study}.

\section{Conclusion}
In this paper, we present STRIDE-ED, a structured empathetic dialogue framework that explicitly models stepwise cognitive reasoning over situation, emotion, and strategy. To support effective learning, we introduce a consistency-based scoring mechanism and a strategy-aware sampling scheme to construct supervision. Extensive experiments demonstrate that STRIDE-ED achieves improvements in empathetic controllability, response relevance, and lexical diversity across multiple settings. These findings suggest that incorporating explicit cognitive structure and data refinement is a promising direction for empathetic dialogue systems.

\section*{Limitations}
Our method currently lacks rigorous statistical validation in terms of data filtering and sampling. We plan to further investigate the underlying statistical principles and mechanisms to improve the soundness and rigor of the method design.
In addition, our experiments indicate that training with DeepSeek-7B-Chat yields the best results; however, we have not yet examined whether this is related to the use of DeepSeek-R1 for data annotation. Furthermore, due to the high computational and time costs, we did not perform extensive hyperparameter tuning on other models. Future work will explore the performance limits of individual backbone models more thoroughly.

\section*{Ethical considerations}
LLMs are used for automatic annotation under prompts enforcing ethical constraints, and they are also applied to the grammatical checking and polishing of texts.  Human evaluators are informed that the evaluated content may contain negative emotions and are compensated fairly for their contributions.  No personal or sensitive information is involved, and all experiments are conducted on publicly available datasets. 

\section*{Acknowledgments}
This research was supported by the National Natural Science Foundation of China (Nos. 62271411, 62471403, 62572403), the Technological Innovation Team of Shaanxi Province (No. 2025RS-CXTD-009), the International Cooperation Project of Shaanxi Province (No. 2025GH-YBXM-017).


\bibliography{custom}

\appendix
\section{Strategy System Details}
\label{sec:appendix_strategy}
The comprehensive strategy system underpinning the STRIDE-ED framework is detailed in Table~\ref{tab:intro_strategy}. It encompasses 14 distinct strategies categorized into three tiers of cognitive and implementational Difficulty (I, II, and III). This taxonomy is constructed to guide the model's reasoning chain across the full emotional spectrum, ranging from fundamental responses like Restatement (Difficulty I) to advanced cognitive interventions such as Cognitive Reframing (Difficulty III). The tiered structure ensures that the framework can adaptively engage in appropriate strategic planning, mirroring the nuanced decision-making process of human empathetic dialogue.

\begin{table*}[htbp]
	\centering	
	\begin{tabular}{
			C{0.001\textwidth}
			C{0.29\textwidth}
			m{0.52\textwidth}
			C{0.08\textwidth}
		}
		\toprule
		\multicolumn{2}{c}{\textbf{Strategy}} & \multicolumn{1}{c}{\textbf{Definition}} & \textbf{Difficulty} \\
		\midrule
		1 & Gratitude Prompting & Encourages the speaker to notice and reflect on positive experiences or supportive aspects of their life, fostering positive emotional awareness. & \uppercase\expandafter{\romannumeral 1} \\ \hline
		2 & Restatement or Paraphrasing & Rephrases the speaker’s main ideas to demonstrate understanding and attentive listening. & \uppercase\expandafter{\romannumeral 1} \\ \hline
		3 & Others & Covers responses that do not clearly fit into any predefined strategy category. & \uppercase\expandafter{\romannumeral 1} \\ \hline
		4 & Information & Provides objective facts or relevant knowledge to help the speaker better understand their situation or make decisions. & \uppercase\expandafter{\romannumeral 1} \\ \hline
		5 & Neutral Validation & Affirms that neutral or low-intensity emotional states are normal and acceptable without encouraging stronger emotions. & \uppercase\expandafter{\romannumeral 2} \\ \hline
		6 & Positive Reinforcement & Highlights the speaker’s strengths, efforts, or constructive behaviors to reinforce confidence and motivation. & \uppercase\expandafter{\romannumeral 2} \\ \hline
		7 & Exploring Actions and Intentions & Uses targeted questions to clarify the speaker’s actions, plans, and underlying intentions. & \uppercase\expandafter{\romannumeral 2} \\ \hline
		8 & Self-disclosure & Shares limited, relevant personal information to foster rapport and mutual understanding. & \uppercase\expandafter{\romannumeral 2} \\ \hline
		9 & Affirmation and Reassurance & Acknowledges the speaker’s feelings and offers comfort or emotional support. & \uppercase\expandafter{\romannumeral 2} \\ \hline
		10 & Reflection of Feelings & Identifies and articulates emotions that the speaker implies but does not explicitly express. & \uppercase\expandafter{\romannumeral 3} \\ \hline
		11 & Cognitive Reframing & Offers an alternative perspective on a difficult situation while respecting the speaker’s original emotions. & \uppercase\expandafter{\romannumeral 3} \\ \hline
		12 & Exploring Feelings and Emotions & Uses open-ended prompts to encourage deeper expression of the speaker’s emotional experience. & \uppercase\expandafter{\romannumeral 3} \\ \hline
		13 & Exploring Thoughts and Cognition & Probes the speaker’s beliefs, interpretations, and thought processes. & \uppercase\expandafter{\romannumeral 3} \\ \hline
		14 & Providing Suggestions & Offers practical and actionable recommendations tailored to the speaker’s needs. & \uppercase\expandafter{\romannumeral 3} \\ 
		\bottomrule
	\end{tabular}
	\caption{The Structured Empathy Strategy System of STRIDE-ED.}
	\label{tab:intro_strategy}
\end{table*}

\section{Prompts}
\label{sec:appendix_prompts}
In the research on conversational strategy analysis and reasoning quality evaluation of large language models, standardized prompts are the key to ensuring the consistency and reliability of task results. Below, we will introduce the design concepts and principles of two core prompts. For the Data Annotation Prompt, it achieves high-quality strategy classification annotation and clarifies the reasons for strategy selection through professional role guidance, closed-set strategy selection, and standardized output. The second one is the Consistency-Based Scoring Prompt, which performs minimal quantitative scoring from a neutral evaluation perspective and completes an objective assessment of the rationality of model reasoning by invoking three different models.

\begin{tcolorbox}[title={Data Annotation Prompt}, colback=white, colframe=black]
	\textbf{Role}: You are an expert in empathetic dialogue and strategy analysis. \\
	\textbf{Inputs:} \\
	1. Context: \{\} \\
	2. Emotion: \{\} \\
	3. Listener's Response: \{\}\\
	\textbf{Task:}
	This is an empathetic conversation, please read the context and focus on the listener's last reply. Now, suppose you are the listener mentioned above—please complete the following tasks: \\
	<Summary> Briefly summarize the speaker's situation. 
\end{tcolorbox}

\begin{tcolorbox}[title={Data Annotation Prompt (Continued)}, colback=white, colframe=black]
	<Strategy> From the first-person perspective, choose one strategy from the optional strategies and their interpretations below and explain how you would apply it, keeping the reasoning concise. \\
	Optional strategies: \\
	1.Exploring Thoughts and Cognition: Probes the speaker’s beliefs, interpretations, and thought processes. \\
	2.Exploring Actions and Intentions: ... \\
	... \\
	14.Others: ... \\
	\textbf{Output Format:} \\
	<Summary> ... \\
	<Strategy> [one strategy], [reason and actions]\\
	\textbf{Output Requirement:} \\
	1. Focus on the speaker's last utterance for the need. \\
	2. Pick only strategies actually used in the listener's response. \\
	3. Be concise and precise.
\end{tcolorbox}

\begin{tcolorbox}[title={Consistency-Based Scoring Prompt},colback=white, colframe=black]
	\textbf{Role}: You are an evaluator. \\
	\textbf{Inputs:} \\
	1.The dialogue context. \{\}\\
	2.The target reply. \{\}\\
	3.The generated reasoning process. \{\}\\
	\textbf{Task:}
	Your task is to assess whether the reasoning process is reasonable and accurate in supporting the target reply, given the dialogue. Please output a single integer score from 1 to 5 (1 = very poor, 5 = excellent).
	\textbf{Output Requirement:}
	Output only the integer score (1-5), without any explanation, extra text, punctuation, or formatting.
\end{tcolorbox}

\section{Experimental Details}
\label{sec:appendix_exp_details}
This appendix section provides an overview of the backbone LLMs used in our experiments, and information regarding the involved existing packages is available in the code repository~\footnote{https://anonymous.4open.science/r/STRIDE-ED/}.

\textbf{Qwen3-0.6B\footnote{https://huggingface.co/Qwen/Qwen3-0.6B}} is a 0.6B-parameter causal language model with 28 layers and a 32,768-token context. It supports thinking and non-thinking modes, demonstrating strong reasoning, agent capabilities, and multilingual instruction following.

\textbf{Qwen3-4B\footnote{https://huggingface.co/Qwen/Qwen3-4B}} is a 4B-parameter causal language model with 36 layers and a a 32,768 natively and 131,072 tokens with YaRN context. It features enhanced multilingual understanding, high-efficiency inference, and robust performance on Chinese and English tasks, excelling in factual accuracy, creative generation, and industrial scenario adaptation.

\textbf{Qwen3-4B-Instruct}\footnote{https://huggingface.co/Qwen/Qwen3-4B-Instruct-2507} is a 4B-parameter causal language model with 36 layers. Optimized for instruction following, it supports zero-shot/few-shot learning, real-time dialogue interaction, and domain-specific task customization, demonstrating superior alignment with human intent and practical application scalability.

\textbf{LLama3.2-3B}\footnote{https://huggingface.co/meta-llama/Llama-3.2-3B} is a 3B-parameter causal language model with 28 layers and a 8,192-token context. It focuses on lightweight deployment, delivering balanced performance in reasoning, code generation, and multilingual processing, with optimized efficiency for edge and low-resource environments.

\textbf{GLM-Z1-9B}\footnote{https://huggingface.co/zai-org/GLM-Z1-9B-0414} is a 9B-parameter causal language model with 40 layers and a 30,000-token context. It integrates self-supervised pre-training with instruction fine-tuning, showcasing strong mathematical reasoning, long-text comprehension, and cross-modal fusion capabilities, tailored for complex industrial and academic reasoning tasks.

\section{Case Study}
\label{sec:appendix_case_study}
In this appendix, we present a set of representative case studies covering higher-difficulty strategies, including Providing Suggestions, Exploring Thoughts and Cognition, Cognitive Reframing, etc., to illustrate the performance of our model. These examples are shown in Tables~\ref{tab:case_study_1} -~\ref{tab:case_study_6}.

Each case provides multi-dimensional core information: dialogue context, high-level empathetic response strategy, model-generated response, and real-scenario reference result. Side-by-side presentation of this information intuitively illustrates the model’s strategy selection logic and linguistic expression ability in complex conversations, and allows direct comparison between model outputs and real results to accurately evaluate the model’s performance in applying high-level empathetic response strategies.

\begin{table}[htbp]
	\centering
	\small % 缩小字体适配内容
	\begin{tabular}{l|p{4cm}} % 第二列用p列实现自动换行
		\toprule
		\textbf{Context} & \textit{Speaker}: I am going to a concert very soon. \\
		\midrule
		\textbf{Summary} & The speaker is anticipating an upcoming concert they will attend.\\
		\textbf{Emotion} & Excited \\
		\textbf{Strategy} & Exploring Actions and Intentions \\
		\hline
		\textbf{Actions} &  I chose Exploring Actions and Intentions because the speaker’s excitement about the concert hints at unfolding plans, and asking about details channels that energy into shared anticipation. \\
		\hline
		\textbf{Generated Response} &That's awesome! Who are you going to see? \\
		\hline
		\textbf{Ground Truth} &  Cool! Who are you going to see? \\
		\bottomrule
	\end{tabular}
	\caption{Case Study~1 illustrating the use of the Cognitive Reframing strategy}
	\label{tab:case_study_1}
\end{table}

\begin{table}[htbp]
	\centering
	\small % 缩小字体适配内容
	\begin{tabular}{l|p{4cm}} % 第二列用p列实现自动换行
		\toprule
		\textbf{Context} & \textit{Speaker}: Speaking of kids in school, I just found out my son got a B in his world history class at the end of the previous school year. I can't believe it. \\
		& \textit{Listener}: That's great!! History is so important I think and so few kids really grasp it well. Congrats to him for a job well done! \\
		& \textit{Speaker}: No no, see, it was the first grade below an A that he's ever made. I honestly thought I had the next Einstein on my hands. Now I guess I'll have to settle for...above average, I guess. :( \\
		
		\midrule
		\textbf{Summary} & The speaker's son achieved a B in history, his first grade below an A, which initially made the speaker hopeful but now feels disappointed.\\
		\textbf{Emotion} & surprised \\
		\textbf{Strategy} & Cognitive Reframing \\
		\hline
		\textbf{Actions} & I chose Cognitive Reframing because the speaker’s disappointment over a single B clashed with sky-high expectations, and recasting the grade as proof of challenging coursework softens the sting and restores perspective. \\
		\hline
		\textbf{Generated Response} & Well, that's still a great grade! I'm sure he'll do better next year. I'm sure you're proud of him! \\
		\hline
		\textbf{Ground Truth} & Oh, I see what you are saying! Did he get upset about it? And you never know, he could turn it around next year and blow you away! \\
		\bottomrule
	\end{tabular}
	\caption{Case Study~2 illustrating the use of the Cognitive Reframing strategy}
	\label{tab:case_study_2}
\end{table}

\begin{table}[htbp]
	\centering
	\small % 缩小字体适配内容
	\begin{tabular}{l|p{4cm}} % 第二列用p列实现自动换行
		\toprule
		\textbf{Context} & \textit{Speaker}: Every year when the kids go back to school, it brings back so many memories of when my kids were going to school. \\
		& \textit{Listener}: I know what you mean. I love to see all the pictures, and think back to how little mine used to be. \\
		& \textit{Speaker}: I would cry the first day of school every year, especially the milestone years. Now my grandchildren are starting school. Time flies by too quickly \\
		
		\midrule
		\textbf{Summary} & The speaker reflects on their children's school years, now experiencing the same bittersweet emotions as their grandchildren begin school, expressing how quickly time passes.\\
		\textbf{Emotion} & Sentimental \\
		\textbf{Strategy} & Self-disclosure \\
		\hline
		\textbf{Actions} &   I chose Self-disclosure because the speaker’s nostalgia over grandchildren starting school echoes my own disbelief at how fast kids grow, and sharing that sense of fleeting time affirms we’re in the same bittersweet boat. \\
		\hline
		\textbf{Generated Response} & I know. I have a 10 year old and a 13 year old, and I can't believe how fast it has gone by. I'm sure you feel the same way. \\
		\hline
		\textbf{Ground Truth} & They do seem to be grown before you know it. Make you wonder how if goes so fast! \\
		\bottomrule
	\end{tabular}
	\caption{Case Study~3 illustrating the use of the Self-disclosure strategy}
	\label{tab:case_study_3}
\end{table}

\begin{table}[htbp]
	\centering
	\small % 缩小字体适配内容
	\begin{tabular}{l|p{4cm}} % 第二列用p列实现自动换行
		\toprule
		\textbf{Context} & \textit{Speaker}: I am going to my local amusement park tomorrow and feeling a certain way about it! \\
		\midrule
		\textbf{Summary} & The speaker is anticipating a visit to an amusement park and feels a mix of excitement and anxiety about the experience.\\
		\textbf{Emotion} & Apprehensive \\
		\textbf{Strategy} & Exploring Thoughts and Cognition \\
		\hline
		\textbf{Actions} &  I chose Exploring Thoughts and Cognition because the speaker’s mixed excitement-and-apprehension about the amusement park suggests hidden worries, and probing the thoughts behind the nerves can turn anxiety into eager curiosity. \\
		\hline
		\textbf{Generated Response} &That sounds like a lot of fun! Are you nervous? \\
		\hline
		\textbf{Ground Truth} & Which way are you feeling? Are you a thrill seeker? \\
		\bottomrule
	\end{tabular}
	\caption{Case Study~4 illustrating the use of the Exploring Thoughts and Cognition strategy}
	\label{tab:case_study_4}
\end{table}

\begin{table}[htbp]
	\centering
	\small
	\begin{tabular}{l|p{4cm}}
		\toprule
		\textbf{Context} & \textit{Speaker}: I'm so happy summer classes are officially over and I'll be able to relax for a week! \\
		& \textit{Listener}: Great what will you be doing during your time off ? \\
		& \textit{Speaker}: I'm not even sure what to do with all this time! I guess I should relax or have some leisure time, but everything in me says to try and be productive even on break. It's that school mentality. \\
		\midrule
		\textbf{Summary} & The speaker is conflicted about balancing relaxation and productivity during their week-long break, struggling with the lingering mindset of being productive despite wanting downtime. \\
		\textbf{Emotion} & Joyful \\
		\textbf{Strategy} & Providing Suggestions \\
		\hline
		\textbf{Actions} &  I chose Providing Suggestions because the speaker felt torn between rest and productivity, and offering concrete ways to enjoy downtime eases the guilt while keeping the momentum alive. \\
		\hline
		\textbf{Generated Response} &I think you should just relax and enjoy the time off. You can always get back to work after that. \\
		\hline
		\textbf{Ground Truth} & I agree with you,you'll feel good about yourself being productive ! \\
		\bottomrule
	\end{tabular}
	\caption{Case Study~5 illustrating the use of the Providing Suggestions strategy.}
	\label{tab:case_study_5}
\end{table}

\begin{table}[htbp]
	\centering
	\small % 缩小字体适配内容
	\begin{tabular}{l|p{4cm}} % 第二列用p列实现自动换行
		\toprule
		\textbf{Context} 
		& \textit{Speaker}: The litter of puppies had been born!\\
		& \textit{Listener}: Nice, amazing to witness birth.\\
		& \textit{Speaker}: All of the puppies survived, truly a lucky amazing occasion!\\
		\midrule
		\textbf{Summary} &  The speaker is celebrating the successful birth of their puppies and expressing gratitude for their survival.\\
		\textbf{Emotion} & joyful \\
		\textbf{Strategy} & Affirmation and Reassurance \\
		\hline
		\textbf{Actions} &  I chose Affirmation and Reassurance because the speaker is deeply joyful about their puppies' birth and expressing thanks for their survival; this reinforces their positive feelings and validates their relief by acknowledging the success and wishing them well. \\
		\hline
		\textbf{Generated Response} &That is a great news, congrats. \\
		\hline
		\textbf{Ground Truth} & Good to hear, have fun with puppies! \\
		\bottomrule
	\end{tabular}
	\caption{Case Study~6 illustrating the use of the Affirmation and Reassurance strategy}
	\label{tab:case_study_6}
\end{table}

\end{document}